\documentclass{article}
\pdfoutput=1
\usepackage[utf8]{inputenc} 
\usepackage[T1]{fontenc}    
\usepackage[numbers]{natbib}
\usepackage{mathrsfs} 
\usepackage{amsmath,amsbsy,amsfonts,amssymb,amsthm,dsfont,units}
\usepackage{mathtools}
\usepackage{color,cases}
\usepackage{tikz}
\usetikzlibrary{calc,shapes}
\usepackage{caption}
\usepackage{subcaption}
\usepackage[utf8]{inputenc}
\usepackage[english]{babel}
\usepackage{multirow}
\usepackage{bbm}
\usepackage{graphicx}
\usepackage{fullpage}
\usepackage{xr}

\usepackage{algorithm}
\usepackage{algpseudocode}
\algnewcommand\algorithmicinput{\textbf{Input: }}
\algnewcommand\INPUT{\State\algorithmicinput}
\algnewcommand\algorithmicinitialize{\textbf{Initialize: }}
\algnewcommand\INIT{\State\algorithmicinitialize}
\algnewcommand\algorithmicrun{\textbf{Run: }}
\algnewcommand\RUN{\State\algorithmicrun}
\algnewcommand\algorithmicupdate{\textbf{Update: }}
\algnewcommand\UPDATE{\State\algorithmicupdate}
\algnewcommand\algorithmicset{\textbf{Set: }}
\algnewcommand\SET{\State\algorithmicset}
\algnewcommand\algorithmicquery{\textbf{Query: }}
\algnewcommand\QUERY{\State\algorithmicquery}
\algnewcommand\algorithmicoutput{\textbf{Output: }}
\algnewcommand\OUTPUT{\State\algorithmicoutput}

\def\norm#1{\mathopen\| #1 \mathclose\|}

\newcommand{\ignore}[1]{}

\def\trace{{\bf Tr}}
\def\trace{{\bf diag}}

\def\reals{{\mathbb R}}

\def\bold0{\mathbf{0}}

\def\bx{\mathbf{x}}

\def\bI{\mathbf{I}}

\def\trace{{\bf Tr}}

\newcommand{\indexI}{\ensuremath{\mathcal I}}

\def\bx{\mathbf{x}}

\def\bI{\mathbf{I}}

\def\eps{\varepsilon}
\def\epsilon{\varepsilon}

\newcommand{\braces}[1]{\left\{#1\right\}}

\DeclareMathOperator{\argmin}{argmin}

\newtheorem{theorem}{Theorem}[section]
\newtheorem{corollary}[theorem]{Corollary}
\newtheorem{definition}[theorem]{Definition}
\newtheorem{lemma}[theorem]{Lemma}

\theoremstyle{definition}

\newcommand\R{\mathbb{R}}

\DeclareMathOperator{\diag}{diag}

\title{Extreme Tensoring for Low-Memory Preconditioning}
\author{
  Xinyi Chen$^1$ \qquad Naman Agarwal$^1$  \\
  Elad Hazan$^{1\,2}$ \qquad Cyril Zhang$^{1\,2}$ \qquad Yi Zhang$^{1\,2}$ \\
  \\
  $^1$ Google AI Princeton \\
  $^2$ Department of Computer Science, Princeton University \\ \\
  \texttt{ \{xinyic,namanagarwal\}@google.com }, \\
  \texttt{\{ehazan,cyril.zhang,y.zhang\}@cs.princeton.edu}
}

\begin{document}
\maketitle

\begin{abstract}
State-of-the-art models are now trained with billions of parameters, reaching hardware limits in terms of memory consumption. This has created a recent demand for memory-efficient optimizers. To this end, we investigate the limits and performance tradeoffs of memory-efficient adaptively preconditioned gradient methods. We propose \emph{extreme tensoring} for high-dimensional stochastic optimization, showing that an optimizer needs very little memory to benefit from adaptive preconditioning. Our technique applies to arbitrary models (not necessarily with tensor-shaped parameters), and is accompanied by regret and convergence guarantees, which shed light on the tradeoffs between preconditioner quality and expressivity. On a large-scale NLP model, we reduce the optimizer memory overhead by three orders of magnitude, without degrading performance.
\end{abstract}


\section{Introduction}

Among the most influential and important optimization techniques in machine learning are adaptive learning-rate methods, otherwise known as diagonal-matrix adaptive preconditioning. Essentially all of the most-commonly used incarnations of adaptive preconditioning (AdaGrad, Adam, RMSprop, Adadelta, etc.) accumulate second-moment estimators of each coordinate of the gradient, then scale the parameter updates by the square roots of these accumulators. These methods come with an overhead memory cost of storing these accumulators, thereby doubling the memory consumption. In the regime where model size encroaches upon the same order of magnitude as the total amount of RAM, a need has arisen to view memory as a limited resource in large-scale optimization.

We address the question of whether the benefits of adaptive preconditioning can be attained without significant memory overhead. To this end, we introduce \emph{extreme tensoring}, a family of generic modifications to any second-moment-based adaptive optimizer. 
Our method uses a compressed preconditioner which takes the form of a tensor product of arbitrary order, with simple updates, without necessarily taking the view of requiring tensor-shaped parameters. In our regret analysis, we quantify how extreme tensoring competes provably with full-memory AdaGrad in the online convex optimization framework, with a multiplicative data-dependent constant that can be measured empirically.

In a large-scale language modeling setting, we demonstrate that an optimizer requires very little additional memory to benefit from adaptive preconditioning. Furthermore, the inherent flexibility of our method enables us to conduct,  to the first of our knowledge,  the first empirical study of the tradeoff between training convergence and memory \emph{in the optimizer}.


\subsection{Related work}

The most widely-adopted form of adaptive preconditioning are second-moment-based: for example, AdaGrad \citep{duchi2011adaptive}, Adam \citep{kingma2014adam}, RMSprop \citep{ tieleman2012lecture}, and Adadelta \citep{zeiler2012adadelta}. 
Some recent preconditioning methods are not based on second moments \cite{bello2017neural,chen2018closing,bernstein2018signsgd}, and fall beyond our scope.



\textbf{Tensor-factorized preconditioners in deep learning.}  \citep{martens2015optimizing,gupta2018shampoo,martens2018kronecker} have investigated tensor-factorized preconditioners. These are presented in the view of restricted full-matrix preconditioning (vs. diagonal for us) of tensor parameters (vs. general-purpose for us). In parameter count regimes relevant to this paper, these full-matrix adaptive methods suffer from prohibitive time and memory overhead issues. The theoretical part of our work follows proof techniques seen in \cite{gupta2017unified} and \cite{gupta2018shampoo}, although our diagonal restriction results in distinct updates and incomparable regret bounds.

\textbf{Memory reduction in the optimizer.} Perhaps the most closely related work to ours is Adafactor \cite{shazeer2018adafactor}, an empirical work which achieves sublinear-memory adaptive regularization by restricting preconditioners on matrix-shaped gradients to ``row'' and ``column'' learning rates, with a similar update rule. Similar algorithms have appeared in prior work \citep{gupta2014training,shazeer2017outrageously}.
As a special case of our regret analysis, we provide some theory for this line of work, while proposing a more general and versatile memory-reduction method.

\paragraph{Our contributions:}
\begin{itemize}
\item We propose extreme tensoring as a modification to AdaGrad, Adam, etc. for reducing the overhead memory cost in adaptive preconditioning.
\item Using the user-selectable degree of memory reduction, we conduct an empirical study of training convergence vs. preconditioner memory usage.
\item We derive a regret bound for extreme-tensored AdaGrad. Strikingly, it competes with AdaGrad within a  multiplicative constant, which we measure to be small in practice.
\end{itemize}

\section{Preliminaries}
\subsection{Stochastic optimization and adaptive methods}
\label{subsec:prelim-setting}

We will state our algorithmic contribution in the usual framework of stochastic optimization of a differentiable function $f(\cdot)$,
equipped with an unbiased stochastic gradient oracle $\widetilde{\nabla} f (\cdot)$. At each of $T$ iterations, the algorithm queries a stochastic gradient $g_t := \widetilde{\nabla} f (x_t)$. In practice, our target is the large-scale ERM setting, where $f$ is the population loss on training data, and $\widetilde{\nabla} \! f$ is a mini-batch stochastic gradient with respect to the model parameters.

However, since the introduction of AdaGrad \citep{duchi2011adaptive}, the standard analysis of adaptive regularization has been in the online convex optimization (OCO) framework (see, e.g. \cite{gupta2017unified,gupta2018shampoo}).
In this language, we view the $x_t$ queried by the learning algorithm as a sequence of online \emph{decisions} chosen from a convex set $\mathcal{K}$, and the $g_t$ as a sequence of gradients of adversarially chosen convex loss functions $f_t$. We are interested in minimizing the regret, defined as
\[ \sum_{t=1}^T f_t(x_t) - \min_{x \in \mathcal{K}} \sum_{t=1}^T f_t(x). \]
This is a generalization of stochastic convex optimization; for a broad survey, see \citep{hazan2016introduction}. A framework for reducing the stochastic \emph{non-convex} setting to the online convex setting is introduced in \citep{agarwal2018case}, translating a sublinear regret bound to convergence to approximate stationary points.

\subsection{Tensors and indexing conventions}
We introduce some notation and terminology for clarity and convenience.

By $[n]$ we denote the index set $\{1, \ldots, n\}$. $\bI_d$ and $\mathbf{0}_d$ refer to the $d\times d$ identity matrix and $d$-dimensional zero vector, respectively.

As is standard in optimization contexts, a tensor of order $p$ is an element of
$\R^{d_1 \times \ldots \times d_p}$:
a $p$-dimensional table of real numbers, indexed by a $p$-tuple $I = (I_1, \cdots I_p)$
of integers, where $I_i \in [d_i]$.
Our methods will require the user to specify a relabeling of gradient vectors, which we will call
a \emph{tensor index}:
\begin{definition}
Let $d_1, \ldots, d_p$ be positive integers whose product is $d$.
A \emph{tensor index} is a bijection $\indexI : [d] \rightarrow \bigtimes_{i=1}^p [d_i]$
between indices for $\R^d$ and $\R^{d_1 \times \ldots \times d_p}$.
\end{definition}
We will refer to the ``reshape'' conversion between vectors $x \in \R^d$ and tensors $\bx \in \R^{d_1 \times \ldots \times d_p}$
specified by the index bijection $\indexI$. For this, we will use the shorthand notation
$\bx := \indexI(x)$ and $x := \indexI^{-1}(\bx)$.
Throughout this paper, we will use square brackets to refer to vector and tensor indices: for example, the previous definition can be restated as enforcing $x[i] = \bx[\indexI(i)]$.

Although we do not need them to state the simple algorithm, it will be useful in the analysis to introduce a few other pieces of notation. We denote the tensor (or Kronecker) product of matrices $A,B$ by $A \otimes B$. Given a positive definite matrix $A \in \reals^{d \times d}$, for any $x \in \reals^{d}$, define the matrix norm $\|x\|_A := \sqrt{ x^\top A x }$ and its dual $\|x\|_A^* := \sqrt{ x^\top A^{-1} x }$. For a square matrix $M$, $\mathrm{diag}(M)$ refers to the diagonal matrix of the same dimensions and diagonal entries as $M$.

\section{Preconditioning by extreme tensoring}
We begin by presenting our proposal for black-box memory reduction in adaptive preconditioning. 
We call this \emph{extreme tensoring}, as it takes the view of restricting a high-dimensional preconditioner to an arbitrary-order tensor product.

The algorithm implicitly maintains a preconditioner which is a rank-one tensor product of the same dimension as a given tensor index. A second-moment accumulator is maintained for sums of squared gradient magnitudes across each $(p-1)$-dimensional tensor slice; the adaptive step size for each coordinate is the inverse square root of the geometric mean of its $p$ corresponding slice sums. The formal specification of the base algorithm (AdaGrad with extreme tensoring) is given in Algorithm~\ref{alg:adagrad-tensor}. 

\begin{algorithm}
\caption{ AdaGrad with extreme tensoring }
\begin{algorithmic}[1]
\INPUT Initializer $x_1$, learning rate schedule $\braces{\eta_t}$, tensor index $\indexI$
with dimensions $(d_1, \ldots, d_p)$, $\eps > 0$
\State Initialize $(S^{(1)}, \ldots, S^{(p)}) := (\mathbf{0}_{d_1}, \ldots, \mathbf{0}_{d_p})$
\For{$t = 1, \ldots, T$}
  \State Receive stochastic gradient $g_t$
  \State Reshape: $\mathbf{g}_t := \indexI(g_t)$
  \State Accumulate slice sums: $$\forall i \in [p], \quad S^{(i)}[j] \leftarrow S^{(i)}[j] + \sum_{I: I_i = j} \mathbf{g}_t[I]^2 $$
  \State Get step sizes: $\delta_t[I] := (\eps + \prod_{i=1}^p S^{(i)}[I_i])^{-\frac{1}{2p}}$
  \State Update: $x_{t+1} \leftarrow x_t - \eta_t \cdot \indexI^{-1}( \delta_t ) \cdot g_t $
\EndFor
\end{algorithmic}
\label{alg:adagrad-tensor}
\end{algorithm}

We make a few important remarks:
\begin{itemize}
\item AdaGrad is a special case of Algorithm~\ref{alg:adagrad-tensor}, with $p=1, d_1 = d$. The analogues for Adam, RMSprop, Adadelta, etc. are obtained straightforwardly by decaying the accumulator ($S^{(i)}[j] \leftarrow \beta_2 \cdot S^{(i)}[j] + \ldots$).
Extreme tensoring is compatible with first-moment estimation (i.e. momentum), although the memory savings disappear. In the setting for our main empirical result, removing momentum did not degrade performance, corroborating the findings of \cite{shazeer2018adafactor}.
\item If the tensor dimensions $(d_1, \ldots, d_p)$ in the decomposition are close to equal, the memory overhead scales as $O( p \cdot d^{1/p} )$.
\item As is standard in matrix and tensor optimization methods \citep{gupta2018shampoo,shazeer2018adafactor,martens2018kronecker}, independent copies of Algorithm~\ref{alg:adagrad-tensor} can be run on each tensor-shaped parameter group; optimizer APIs in standard deep learning packages promote this convention. This can be viewed as maintaining a tensor sum of preconditioners, each of which is a $p$-wise tensor product.
\item The accumulator update step can be implemented concisely and efficiently using re-indexing procedures (typically called \texttt{reshape} or \texttt{view}) in standard deep learning and numerical linear algebra packages. 
\end{itemize}

Lemma~\ref{lem:overestimate} in the analysis shows that the per-coordinate learning rates are \emph{underestimates} of those prescribed by AdaGrad. This interpretation is used in our proof, and serves as the basis of the empirical measurements in Section~\ref{subsec:regret-measurement}.

\section{Regret analysis}
In this section, we justify the choice of update rules for the compressed preconditioner using the regret-minimizing adaptive regularization framework introduced in the original analysis of AdaGrad \citep{duchi2011adaptive}. Throughout the analysis, we will adopt the conventions of online convex optimization, as discussed in Section~\ref{subsec:prelim-setting}. This gives rise to an intuitively appealing and interpretable regret bound, which we discuss in this section and revisit in the empirical measurements of Section~\ref{subsec:regret-measurement}.

\subsection{Statement and discussion of regret bound}
First, we define and prove the main regret bound. As is standard in the regret analysis of diagonal AdaGrad (see \cite{duchi2011adaptive}), this is most naturally stated with a diameter term on $x_t$: $D_\infty := \max_{t, x^*} \norm{x_t - x^*}_{\infty}$. This can be constrained using a projection onto an $\ell_\infty$ ball in the algorithm, but this is seldom seen in practice. It will also be helpful to define $\mathbf{G}_t^i$ to be the diagonal matrix with values
$$
\mathbf{G}_t^i[j,j]  = \sum_{I: I_i=j} \mathbf{g}_t[I]^2
$$ on the diagonal.

\begin{theorem}
\label{thm:main}
Define $H_T = \otimes_{i=1}^p (\epsilon \bI_{d_{i}} + \sum_{t=1}^T \mathbf{G}_t^{i})^{1/2p}$, and $\hat{H}_T = \diag(\epsilon \bI + \sum_{t=1}^T g_tg_t^\top)^{1/2}$. Then, there exists a choice of constant learning rate schedule $\eta_1 = \ldots = \eta_t := \eta$ and $\eps > 0$ such that the $\{x_t\}$ chosen by Algorithm 1 satisfy the regret bound 
$$
\sum_{t=1}^T f_t(x_t) - f_t(x^*) \le D_\infty \sqrt{2\trace(H_T)\trace(\hat{H}_T)},
$$
where
$$x^* := \min_x \sum_{t=1}^T f_t(x)$$
is the loss-minimizing decision in hindsight.
\end{theorem}

As a direct consequence, we recover the AdaGrad regret bound as a special case where $p=1$. Noting further that $H_T = \hat H_T$, we restate this well-known result in our notation, for clarity and comparisons:
\begin{corollary}
\label{cor:adagrad}
In setting of Theorem~\ref{thm:main}, when $p=1$, the $\{x_t\}$ chosen by Algorithm 1 satisfy
$$
\sum_{t=1}^T f_t(x_t) - f_t(x^*) \le \sqrt{2}D_\infty \trace(\hat H_T).
$$
\end{corollary}

Thus, AdaGrad with extreme tensoring satisfies a regret bound at most $\sqrt{ \trace(H_T) / \trace(\hat{H}_T) }$ times than that of AdaGrad. This quantity can be as large as $\Omega(\sqrt{d})$; this is the worst-case price of the dramatic memory savings in the regularizer. However, in the presence of sparsity, this ratio can be much smaller; we include a discussion and empirical measurements in Section~\ref{subsec:regret-measurement}.

\subsection{Proofs}
The derivation of the regret bound uses the proof framework introduced by \cite{gupta2017unified}. Our first important lemma states that the regularizer in Algorithm~\ref{alg:adagrad-tensor} is a spectral (per-coordinate) upper bound for the diagonal AdaGrad preconditioner. In other words, this lemma establishes that the per-coordinate learning rates under extreme tensoring are \emph{underestimates} of those dictated by AdaGrad.
\begin{lemma} \label{lem:overestimate}
Suppose $\mathbf{g}_1, \mathbf{g}_2, \cdots, \mathbf{g}_T$ are tensors of dimension $d_1 \times \cdots \times d_p$, and let $g_t = \indexI^{-1}(\mathbf{g}_t)$ for all $t$,
where $\indexI$ is any tensor index whose image is $\bigtimes_{i=1}^p [d_i]$. Then for $t\in[T]$, 
$$
\diag(\epsilon \bI + \sum_{s=1}^t g_sg_s^\top)^{1/2} \preceq  \otimes_{i=1}^p (\epsilon \bI_{d_{i}} + \sum_{s=1}^t \mathbf{G}_s^{i})^{1/2p}.
$$
\end{lemma}
\begin{proof}
Let $d = d_1 d_2 \cdots d_p$. Let $j \in [d]$ and let $a_1, \ldots, a_p$ be such that $\mathcal{I}(j) = [a_1, a_2, \cdots, a_p]$.
\begin{align*}
\diag(\epsilon \bI &+ \sum_{s=1}^t g_sg_s^\top)[j,j]^p = (\epsilon + \sum_{s=1}^t g_s[j]^2)^p \\
&= (\epsilon + \sum_{s=1}^t \mathbf{g}_s[a_1, a_2, \cdots, a_p]^2)^p \\
&\le \Pi_{i=1}^p (\epsilon  + \sum_{s=1}^t \sum_{I: I_i = a_i} \mathbf{g}_s[I]^2).
\end{align*}
Taking the $p$th root on both sides,  
\begin{align*}
\diag(\epsilon \bI &+ \sum_{s=1}^t g_sg_s^\top)[j,j] \le \Pi_{i=1}^p (\epsilon  + \sum_{s=1}^t \sum_{I: I_i = a_i} \mathbf{g}_s[I]^2) ^{1/p} \\
&= \Pi_{i=1}^p (\epsilon  + \sum_{s=1}^t\mathbf{G}_s^i[a_i,a_i]) ^{1/p} \\
&= \Pi_{i=1}^p (\epsilon \bI_{d_i}  + \sum_{s=1}^t\mathbf{G}_s^i)[a_i,a_i] ^{1/p} \\
&= \otimes_{i=1}^p (\epsilon \bI_{d_{i}} + \sum_{s=1}^t \mathbf{G}_s^{i})^{1/p}[j,j].
\end{align*}
Taking square roots of the above inequality yields the lemma.
\end{proof}
\subsection{Regret bound}
The following lemma bounds the regret in terms of the quadratic norms in the regularizer and its dual:

\begin{lemma}\label{regret1}
Denote by $H_t$ the $d$-by-$d$ diagonal matrix whose entries are $(\mathcal{I}^{-1}(\delta_t))^{-1}$. Take $\eta_1 = \ldots = \eta_t := \eta$ to be the constant learning rate, and let $x^*$ be the loss minimizing choice in hindsight. Then, the regret of Algorithm 1 is bounded by 
\begin{align*}
\frac{1}{2\eta} \sum_{t=1}^T (\|x_t - x^*\|_{H_t}^2 &- \|x_{t+1} - x^*\|_{H_t}^2) \\
&+ \frac{\eta}{2} \sum_{t=1}^T (\|g_t\|_{H_t}^*)^2.
\end{align*}
\end{lemma}
\begin{proof}
By the definition of Algorithm 1, for any $x^*$,
$$
x_{t+1} - x^* = x_t - x^* - \eta H_t^{-1}g_t,\ \text{and}
$$
$$
H_t(x_{t+1} - x^*) = H_t(x_t - x^*) - \eta g_t
$$
Taking the inner product of the above vectors,
\begin{align*}
& (x_{t+1} - x^*)^\top H_t(x_{t+1} - x^*) \\
&= (x_t - x^*)^\top H_t (x_t - x^*) \\ &\qquad - 2\eta(x_t - x^*)^\top g_t + \eta^2 g_t^\top H_t^{-1}g_t.
\end{align*}
Rearranging and dividing by $2\eta$, we have
\begin{align*}
g_t^\top (x_t - x^*) = \frac{1}{2\eta}(\|x_{t} - x^*\|_{H_t}^2 &- \|x_{t+1} - x^*\|_{H_t}^2) \\ &+ \frac{\eta}{2}(\|g_t\|_{H_t}^*)^2.
\end{align*}
Using convexity of the loss functions and sum over $t$, we finally bound the regret as follows:
\begin{align*}
&\sum_{t=1}^T f_t(x_t) - f_t(x^*) \le \sum_{t=1}^T g_t\top (x_t - x^*) \\
& \le \frac{1}{2\eta} \sum_{t=1}^T (\|x_{t} - x^*\|_{H_t}^2 - \|x_{t+1} - x^*\|_{H_t}^2) \\
&+ \frac{\eta}{2} \sum_{t=1}^T(\|g_t\|_{H_t}^*)^2
\end{align*}
\end{proof}
Next, we present a lemma that is later used to bound the second term in the regret. 

\begin{lemma}\label{potential} \cite{gupta2018shampoo}
Let $g_1, g_2, \cdots, g_T$ be a sequence of vectors, and let $M_t = \diag(\sum_{x=1}^t g_sg_s^\top)$. Let $S^{+}$ denote the set of diagonal positive definite matrices. Given a function $\Phi$ over $S^{+}$, define
$$
A_t = \argmin_{A\in S^{+}} \{M_t \bullet A^{-1} + \Phi(A)\}
$$
and assume that the minimum is attained for all $t$. Then 
$$
\sum_{t=1}^T (\|g_t\|_{A_t}^*)^2 \le \sum_{t=1}^T (\|g_t\|_{A_T}^*)^2 + \Phi(A_T) - \Phi(A_0).
$$
\end{lemma}
We proceed to prove Theorem~\ref{thm:main}.
\begin{proof}(Proof of Theorem~\ref{thm:main}) Recall the definition of $H_t = \otimes_{i=1}^p (\epsilon \bI_{d_{i}} + \sum_{s=1}^t \mathbf{G}_s^{i})^{1/2p}$, 
we bound the first term in Lemma~\ref{regret1} as follows,
\begin{align*}
& \sum_{t=1}^T (\|x_t - x^*\|_{H_t}^2 - \|x_{t+1} - x^*\|_{H_t}^2) \\
&\le  \sum_{t=2}^T (x_t - x^*)^T(H_t - H_{t-1})(x_t - x^*) + \|x_1 - x^*\|_{H_1}^2 \\
&\le D_\infty^2 \sum_{t=2}^T \trace(H_t - H_{t-1}) + D_\infty^2  \trace(H_1) = D_\infty^2\trace(H_T)
\end{align*}
The last inequality is due to the fact that $H_{t-1} \preceq H_t$.
Recall $\hat{H}_t = \diag(\epsilon \bI + \sum_{s=1}^t g_sg_s^\top)^{1/2}$, and by Lemma \ref{lem:overestimate}, $\hat{H}_t \preceq H_t$. Now we use Lemma \ref{potential} with function 
$$
\Phi(A) = \trace(A) + \epsilon \trace(A^{-1}).
$$
Let $M_t = \diag(\sum_{s=1}^t g_sg_s^\top)$ and let $S^+
$ denote the set of diagonal positive definite matrices, we have 
\begin{align*}
&\argmin_{A\in S^+} \{M_t\bullet A^{-1} + \Phi(A)\} \\
&= \argmin_{A\in S^+} \trace(\hat{H}_t^2A^{-1} + A) \\
&= \hat{H}_t
\end{align*}
The last equality can be derived by minimizing each diagonal entry of $H$ individually.

By Lemma \ref{potential}, 
\begin{align*}
& \sum_{t=1}^T (\|g_t\|_{\hat{H}_t}^*)^2 \le \sum_{t=1}^T (\|g_t\|_{\hat{H}_T}^*)^2 + \Phi(\hat{H}_T) - \Phi(\hat{H}_0) \\
&\le \sum_{t=1}^T (\|g_t\|_{\hat{H}_T}^*)^2 + \Phi(\hat{H}_T) \\
&= \diag(\sum_{t=1}^T g_tg_t^\top)\bullet \hat{H}_T^{-1} + \trace(\hat{H}_T) + \epsilon\trace(\hat{H}_T^{-1}) \\
&= \diag(\sum_{t=1}^T g_tg_t^\top + \epsilon \bI)\bullet \hat{H}_T^{-1} + \trace(\hat{H}_T) \\
&= \hat{H}_T^{2}\bullet \hat{H}_T^{-1} + \trace(\hat{H}_T) \\
&= 2\trace(\hat{H}_T)
\end{align*}
We can take the diagonal of $\sum_{t=1}^T g_tg_t^\top$ in the first equality since $\hat{H}_T^{-1}$ is a diagonal matrix as well.

We proceed to bound the second term in Lemma \ref{regret1}. Since $\hat{H}_t \preceq H_t$ and both $H_t$ and $\hat{H}_t$ are full-rank, $\hat{H}_t^{-1} \succeq H_t^{-1}$. Therefore
$$
\sum_{t=1}^T(\|g_t\|_{H_t}^*)^2 \le \sum_{t=1}^T(\|g_t\|_{\hat{H}_t}^*)^2 \le 2\trace(\hat{H}_T).
$$
Summing up the two terms and taking $\eta = D_\infty \frac{\sqrt{\trace(H_T)}}{\sqrt{2\trace(\hat{H}_T)}}$, we conclude that the regret of Algorithm 1 is bounded by 
$$
\frac{1}{2\eta}D_\infty^2 \trace(H_T) + \eta \trace(\hat{H}_T) = D_\infty\sqrt{2\trace(H_T)\trace(\hat{H}_T)}.
$$
\end{proof}

\section{Experiments}
In this section, we provide several empirical studies on extreme tensoring. Our main experiment interpolates between the memorylessness of SGD and the full memory consumption of AdaGrad on a large-scale language model; we additionally isolate the effects of preconditioner expressivity in a synthetic experiment of the same form, and provide a parallel CIFAR-10 experiment in the appendix.

To accompany the main experimental result, we provide empirical measurements of the competitive ratio in the regret bound, as well as a quick comparison in which the memory savings are used to train a larger model.

\subsection{Memory-performance tradeoff in large-scale NLP}
Our main empirical study focuses on large-scale language modeling with the Transformer architecture \citep{vaswani2017attention} on the Google Billion Words (GBW) dataset \citep{chelba2013one}. We use the pipeline from the open-source Tensor2Tensor package \cite{tensor2tensor} for preprocessing the GBW dataset, and use the base Transformer architecture in the same repository as our base architecture. The decoder-only model has 6 identical layers with hidden dimension $d_{model} = 512$, and feedforward dimension $d_{ff} = 2048.$ In addition, the weights are shared between the embedding and softmax layers, and the model has a total of $\sim\!35$M parameters.

For our experiments, we use a learning rate schedule of $\eta_t = c \cdot \min(10^{-6} \cdot t, \frac{1}{\sqrt{t}})$, and $c$ is a hyperparameter we tune for each experiment. The learning rate schedule is the same as the one used in \citep{shazeer2018adafactor}: a linear warmup stage, followed by inverse square root decay. We train each model for 500K steps, with a max sequence length of 256 tokens, and a max number of 4096 tokens in a batch. Global learning rates are selected by hyperparameter search.

Except in our comparison with Adam, we do not use momentum in our extreme tensoring experiments. Momentum incurs an overhead memory cost linear in the model dimension. In addition, we empirically tested the benefit of exponentially decaying the second moment estimator ($\beta_2 < 1$ in the language of Adam and Adafactor). We found that in language modeling experiments, decaying the second moment estimator did not contribute to better performance; on the other hand, the vision experiments in the appendix do use this decay ($\beta_2 = 0.99$).

Extreme tensoring gives us a class of optimizers which interpolate smoothly between AdaGrad and SGD. Between these endpoints, we choose three levels of extreme tensoring (denoted by ET$\{1,2,3\}$), with tensor index dimensions fully specified in the appendix.
Figure~\ref{fig:transformer} records the training performance for these interpolating algorithms. We also provide training results for Adam (which consumes more memory for first-moment estimation) and Adafactor (similar to ET1 but with a different step size scaling) in Table~\ref{table:main}, for completeness; however, these are \emph{not} part of the interpolation study.

To provide an additional interpolating point, we include in the comparison a closely related algorithm, which selects a single learning rate per parameter group (e.g. embedding, bias, convolutional layer); thus, the preconditioner is a tensor sum of scalar multiples of the identity matrix. Each such scalar is chosen to be the inverse square root of the sum of squared $\ell_2$ norms of the parameter group over time; it is easily seen that this achieves the same regret as online gradient descent \cite{zinkevich2003online}, so we include it as the least granular adaptive optimizer, and call it ET$\infty$.

\begin{figure}
  \centering
  \includegraphics[width=0.7\linewidth]{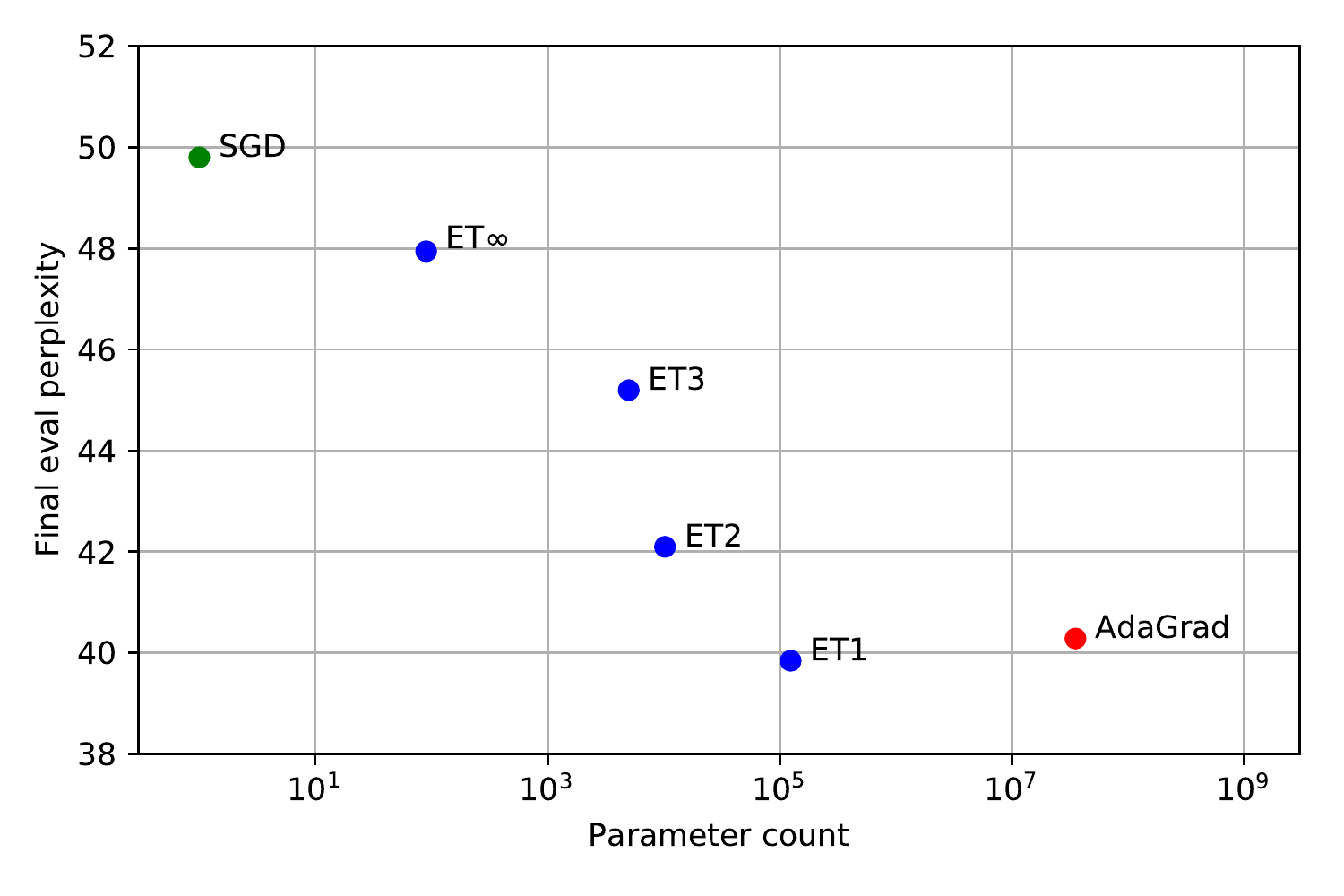}
  \caption{Memory-performance tradeoff for language modeling with a Transformer network: final validation perplexity vs. optimizer parameter count. Note that the horizontal scale is \textbf{logarithmic}, so optimizer memory savings are by orders of magnitude. }
  \label{fig:transformer}
\end{figure}

\begin{table}
\centering
\begin{tabular}{|l|l|l|}
\hline
{\bf Optimizer} & {\bf Parameter count} & {\bf Final ppl.} \\ \hline
AdaGrad & $3.5 \times 10^7$ & 41.18 \\ \hline
ET1 & $1.2 \times 10^5$ & 39.84 \\ \hline
ET2 & $1.0 \times 10^4$ & 42.10 \\ \hline
ET3 & $5.0 \times 10^3$ & 45.19 \\ \hline
ET$\infty$ & 90 & 47.94 \\ \hline
SGD & 1 & 49.80 \\ \hline \hline
Adam & $7.0 \times 10^7$ & 38.47 \\ \hline
Adafactor & $1.2 \times 10^5$ & 38.86 \\ \hline
\end{tabular}
\caption{Performance and memory comparison between adaptive optimizers for GBW language modeling with a Transformer network.}
\label{table:main}
\end{table}

\subsection{Doubling the memory allowance}
As an extension of the above comparison, we argue that the memory consumption freed up by choosing a compressed adaptive optimizer can be usefully reallocated to training a larger model. We double the number of layers of the Transformer network used in the previous section, keeping all embedding sizes the same; this results in a network with 56M parameters.

Results are shown in Table~\ref{table:double-mem}. Holding the memory consumption constant, we find that it pays off to use a larger model rather than an adaptive optimizer that stores all of the second-moment accumulators. Even holding the running time constant, the larger models are competitive with the fully-converged smaller ones.

\begin{table}
\centering
\begin{tabular}{|l|l|l|}
\hline
{\bf Optimizer} & {\bf Final ppl.} \emph{(time)} & {\bf Final ppl.} \emph{(iters)} \\ \hline
ET1 & 39.25 & 36.23 \\ \hline
ET2 & 43.81 & 40.04 \\ \hline
ET3 & 44.70 & 40.45 \\ \hline
ET$\infty$ & 42.95 & 41.68 \\ \hline
\end{tabular}
\caption{Performance comparison of memory-efficient optimizers on a model with doubled size (12-layer Transformer), so that total memory consumption is lower than that of AdaGrad/Adam on the smaller model. Final perplexities are given with the same running time allowance as the corresponding main experiment (middle column), as well as the same iteration count (500K steps; right column).}
\label{table:double-mem}
\end{table}

\subsection{Empirical measurements of the regret bound}
\label{subsec:regret-measurement}
\begin{figure}
  \centering
  \includegraphics[width=0.7\linewidth]{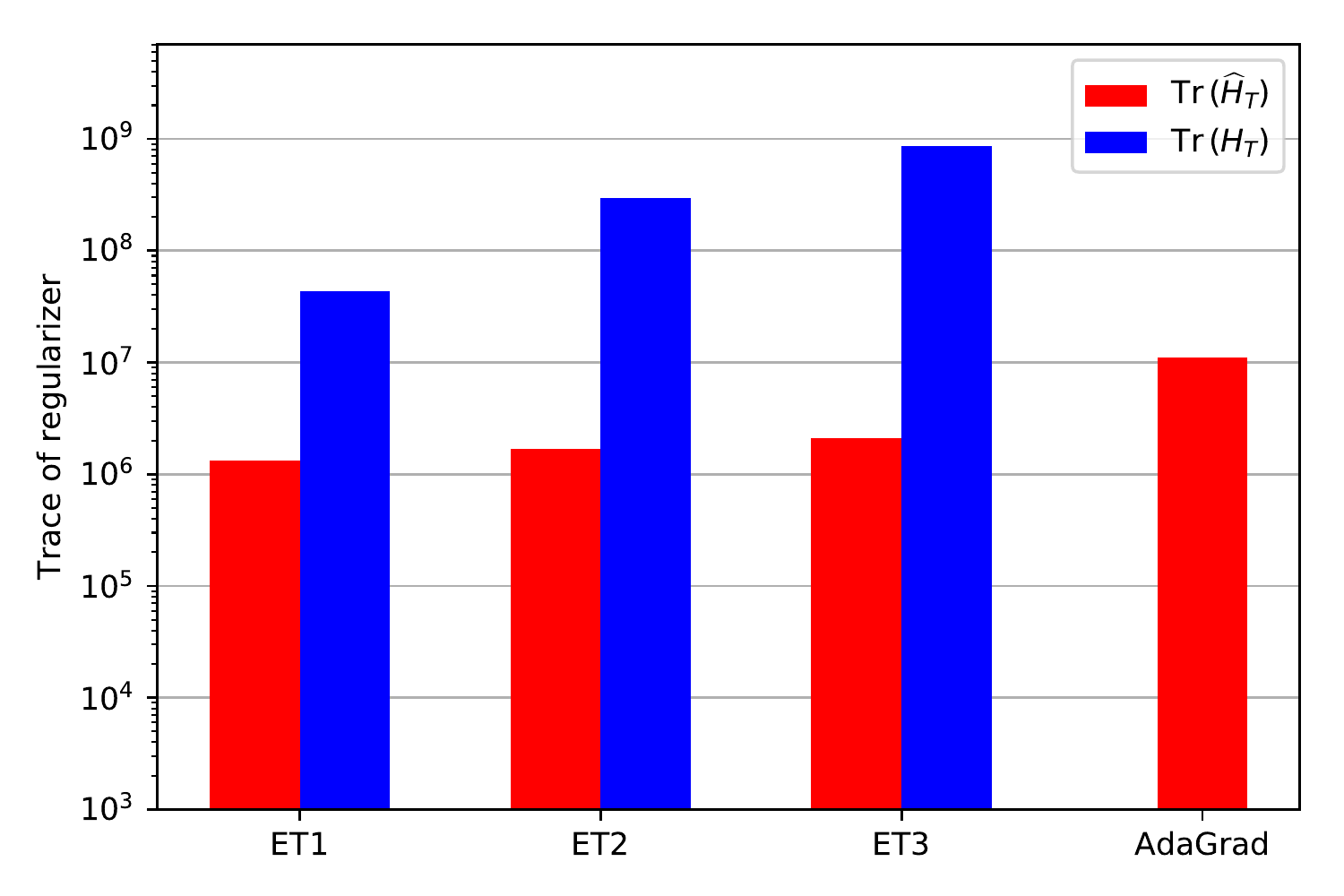}
  \caption{ Comparison of quantities in the numerical regret bounds. Note that the vertical scale is \textbf{logarithmic}; consequently, the multiplicative regret bound gap compared to AdaGrad $\sqrt{ \trace(H_T) / \trace(\hat{H}_T) }$ is visualized as half the height difference between blue and red bars. }
  \label{fig:bounds}
\end{figure}

An appealing part of our regret analysis is the interpretability of the trace quantities in Theorem~\ref{thm:main}. Table~\ref{fig:bounds} shows a comparison of traces $\trace(H_T)$ and $\trace(\hat{H}_T)$ of the final regularizers in the language modeling experiment. Then, by Theorem~\ref{thm:main} and Corollary~\ref{cor:adagrad}, the upper bound for AdaGrad's regret scales with the latter quantity, while that of extreme tensoring scales with the geometric mean of the two traces.

Intriguingly, the multiplicative gap between these two regret bounds, which depends on the loss function and training trajectory via the sequence of gradients, appears to be very small in practice. In the case of one-level extreme tensoring, we have $\sqrt{ \trace(H_T) / \trace(\hat{H}_T) } \approx 5.7$.

\subsection{Comparison on synthetic data}
In this section, we exhibit a simple experiment with a convex optimization problem (logistic regression), in which there is a clear tradeoff between preconditioner quality and expressivity. We generate Gaussian data $x_i \in \R^{512}$ and a Gaussian matrix $W \in \R^{10 \times 512}$. The covariance matrix of $\{x_i\}$ has high condition number ($\sim 10^4$). Labels are generated according to the log-linear model $\Pr[y_i = j] \propto \exp((Wx_i)_j)$. Then, the optimization problem in $W$ is to minimize the empirical negative log-probability under the same model. Our findings are robust with respect to the batch size; we use the full gradient on $10^4$ samples in our plots for clarity.

\begin{figure}
  \centering
  \includegraphics[width=0.7\linewidth]{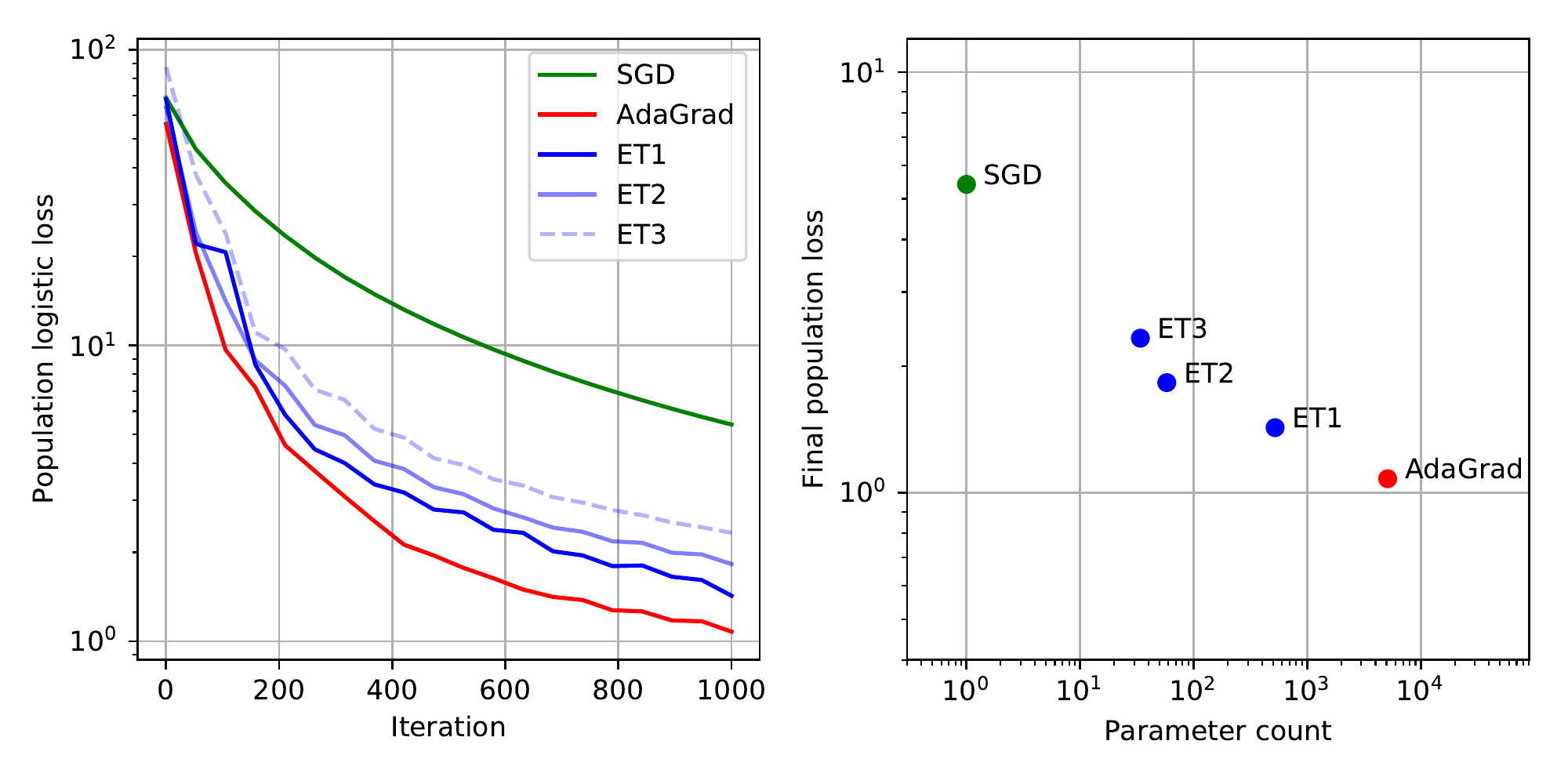}
  \caption{Training curves and final loss comparison for a convex problem with synthetic data.
  \emph{Left:} Population loss for each optimizer vs. iteration number.
  \emph{Right:} Final loss vs. optimizer parameter count. }
  \label{fig:synthetic}
\end{figure}

As in the large-scale experiment, we use successively deeper tensor factorizations of the preconditioner, along the feature dimension of the matrix $W$. For depths 1, 2, and 3, we choose tensor indices of dimensions $(10,512)$, $(10,16,32)$, and $(10,8,8,8)$, respectively. Global learning rates are tuned individually. Results are shown in Figure~\ref{fig:synthetic}.

\section{Conclusion}
We have introduced \emph{extreme tensoring}, a modification to any second-moment-based adaptive optimizer which drastically shrinks memory overhead. Our experiments characterize a performance-memory tradeoff in the optimizer, and demonstrate the possibility of negligible memory overhead without degrading convergence. Our regret analysis provides a competitive ratio with uncompressed adaptive methods, giving an additional empirical lens on data-dependent preconditioner quality.

\bibliography{main}
\bibliographystyle{alpha}

\newpage
\appendix

\section{CIFAR-10 experiment}
In this section, we evaluate the memory-performance trade-off of our proposed algorithm on the CIFAR-10 dataset \citep{krizhevsky2009learning}. Specifically, we compare the test accuracy of a ResNet-18 model \citep{he2016deep} trained with SGD, Adam and 3 levels of extreme tensoring. 

\subsection{Setup}
With each optimizer, the model is trained for 150 epochs with optimally tuned constant learning rate. To prevent a 2X memory overhead, no momentum is used for any of the optimizers (i.e. $\beta_1=0$ in Adam). We use batch size $128$, weight decay $5\times10^{-4}$ for all the experiments in this section.

\subsection{Tensor indices}
We show our tensor decomposition scheme in the table below. Note that bias parameters are not shown in each layer in the table since are treated as vectors and thus not decomposed at all.
\begin{table}[H]
\label{cifar_tensor_indices}
\centering
\begin{tabular}{|l|l|l|l|}
\hline
Parameter Shape       & Shape (ET1)      & Shape (ET2) & Shape (ET3) \\ \hline 
$(64, 3, 3, 3)$           &     $(64, 3, 9)$     &   $(8, 8, 3, 9)$  &   $(8, 8, 3, 9)$  \\ \hline
$(64, 64, 3, 3)$         &    $(64, 64, 9)$   &    $(8, 8, 8, 8, 9)$  &   $(8, 8, 8, 8, 9)$         \\ \hline 
$(128, 64, 3, 3)$    &   $(128, 64, 9)$  &  $(8, 16, 8, 8, 9)$ & $(8, 4, 4, 8, 8, 9)$ \\ \hline 
$(128, 128, 3, 3)$        &  $(128, 128, 9)$  &   $(8, 16, 8, 16, 9)$         & $(8, 4, 4, 8, 4, 4, 9)$ \\ \hline 
$(256, 128, 3, 3)$         &   $(256, 128, 9)$    &  $(16, 16, 8, 16, 9)$    & $(4, 4, 4, 4, 8, 4, 4, 9)$ \\ \hline 
$(256, 256, 3, 3)$          &   $(256, 256, 9)$    &  $(16, 16, 16, 16, 9)$          & $(4, 4, 4, 4, 4, 4, 4, 4, 9)$\\ \hline
$(512, 256, 3, 3)$          &   $(512, 256, 9)$    &  $(32, 16, 16, 16, 9)$          & $(8, 4, 4, 4, 4, 4, 4, 4, 9)$\\ \hline
$(512, 512, 3, 3)$          &   $(512, 512, 9)$    &  $(32, 16, 32, 16, 9)$          & $(8, 4, 4, 4, 8, 4, 4, 4, 9)$\\ \hline
$(128, 64, 1, 1)$      &   $(128, 64)$    &  $(16, 8, 8, 8)$          & $(4, 4, 8, 8, 8)$\\ \hline
$(256, 128, 1, 1)$      &   $(256, 128)$    &  $(16, 16, 16, 8)$       & $(4, 4, 4, 4, 4, 4, 8)$\\ \hline
$(512, 128, 1, 1)$      &   $(512, 128)$    &  $(32, 16, 16, 8)$          & $(8, 4, 4, 4, 4, 4, 8)$\\ \hline
\end{tabular}
\caption{Tensor indices used for different levels of extreme tensoring for the ResNet-18 model on Cifar-10.}
\end{table}

\subsection{Results}
In this section, we report the best test errors seen in the first 150 epoch along with the parameter count in the optimizer. A similar trend of trade-offs between performance and parameter count is observed.

\begin{table}[H]
\centering
\begin{tabular}{|l|l|l|}
\hline
{\bf Optimizer} & {\bf Parameter count} & {\bf Final test error} \\ \hline
Adam($\beta_1=0$) & $1.1 \times 10^7$ & 8.40 \\ \hline
ET1 & $2.3 \times 10^4$ & 7.22 \\ \hline
ET2 & $1.6 \times 10^4$ & 8.49 \\ \hline
ET3 & $1.5 \times 10^4$ & 8.52 \\ \hline
ET$\infty$ & 62 & 8.57 \\ \hline
SGD & 1 & 9.27 \\ \hline
\end{tabular}
\caption{Performance and memory comparison between adaptive optimizers for Cifar-10 classification with a ResNet-18 network.}
\label{table:main}
\end{table}

\begin{figure}[H]
  \centering
  \includegraphics[width=0.6\linewidth]{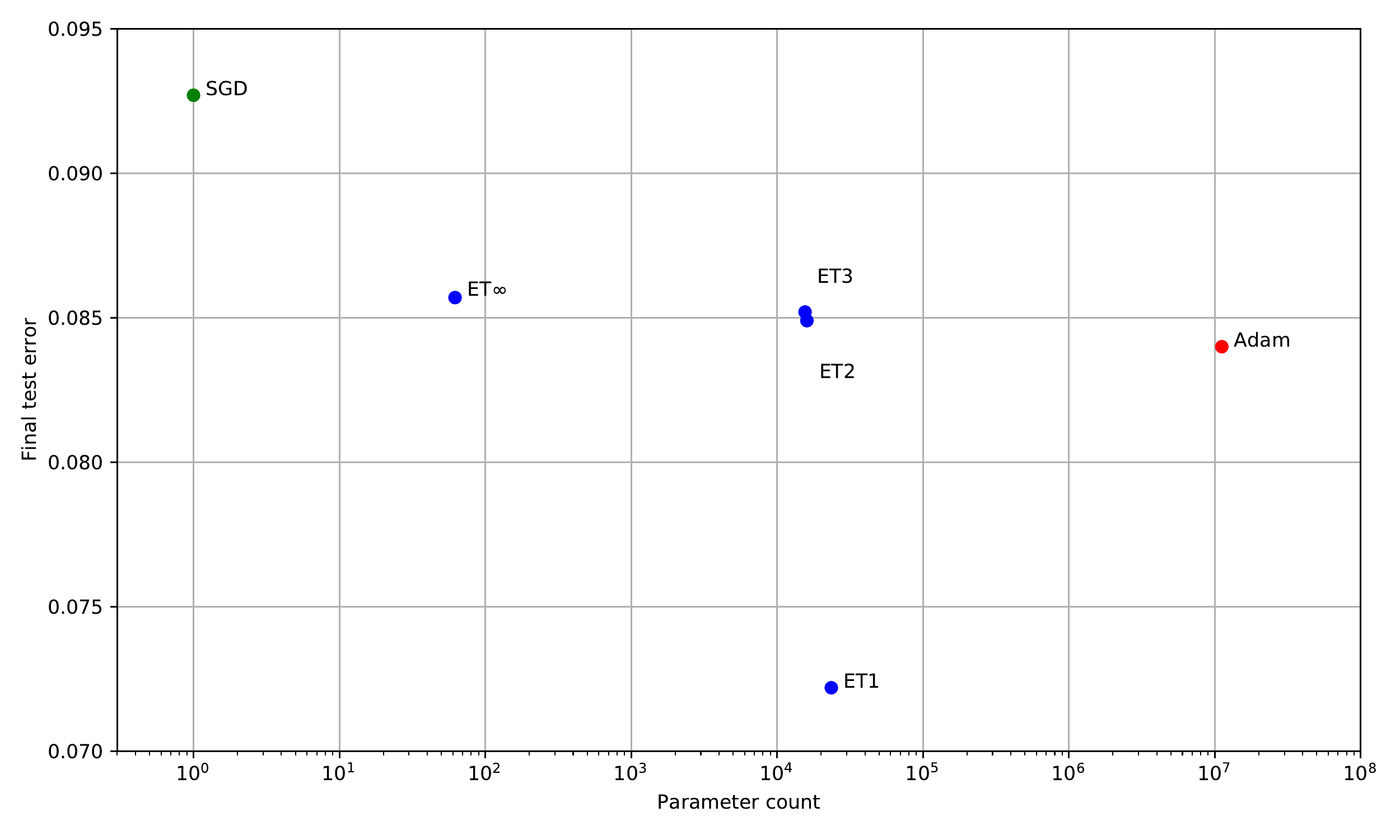}
  \caption{Memory-performance comparison for CIFAR-10 classification with an 18-layer ResNet: final test error vs. optimizer parameter count. Note that the horizontal scale is \textbf{logarithmic}. }
  \label{fig:transformer}
\end{figure}

\section{Language modeling experiment details}
In this section, we provide more details for the main empirical result.

\subsection{Tensor indices}
\begin{table}[H]
\begin{tabular}{|l|l|l|l|l|}
\hline
Parameter & Parameter Shape       & Shape (ET1)      & Shape (ET2) & Shape (ET3) \\ \hline 
$Q,K,V$, Position-wise Feed Forward   & $(512,512)$ & $(512,512)$  & $(16,32,16,32)$           &   $(4, 4, 4, 8, 4, 4, 4, 8)$         \\ \hline
Embedding Weights          &   $(2000,512)$           &     $(2000,512)$            &   $(40, 50, 16, 32)$         &   $(5, 8, 5, 10, 4, 4, 4, 8)$         \\ \hline
Layer Norm          &     $(512,)$         &$(512,)$          &    $(16, 32)$        &   $(4, 4, 4, 8)$         \\ \hline Fully Connected Weights       &    $(512,2048)$          &   $(512,2048)$              &   $(16, 32, 32, 64)$         & $(4, 4, 4, 8, 4, 8, 8, 8)$ \\ \hline Fully Connected Bias
&      $(2048,)$        &         $(2048,)$          &   $(32, 64)$         & $(4, 8, 8, 8)$
\\ \hline Fully Connected Weights
&     $(2048,512)$         &   $(2048,512)$              &        $(32, 64, 16, 32)$    & $(4, 8, 8, 8, 4, 4, 4, 8)$
\\ \hline Fully Connected Bias
&    $(512,)$          &   $(512,)$               &  $(16, 32)$          & $(4, 4, 4, 8)$\\
\hline
\end{tabular}
\end{table}


\end{document}